# Trustworthy Orchestration Artificial Intelligence by the Ten Criteria with Control-Plane Governance

[1]Prof. Byeong Ho Kang       [1]Dr. Wenli Yang       [1]Dr. Muhammad Bilal Amin
[1]School of ICT, College of Science and Technology, University of Tasmania, Australia

## Abstract

As Artificial Intelligence (AI) systems increasingly assume consequential decision-making roles, a widening gap has emerged between technical capabilities and institutional accountability. Ethical guidance alone is insufficient to counter this challenge; it demands architectures that embed governance into the execution fabric of the ecosystem. This paper presents the Ten Criteria for Trustworthy Orchestration AI, a comprehensive assurance framework that integrates human input, semantic coherence, audit and provenance integrity into a unified Control-Panel architecture. Unlike conventional agentic AI initiatives that primarily focus on AI-to-AI coordination, the proposed framework provides an umbrella of governance to the entire AI components, their consumers and human participants. By taking aspiration from international standards and Australia's National Framework for AI Assurance initiative, this work demonstrates that trustworthiness can be systematically incorporated (by engineering) into AI systems, ensuring the execution fabric remains verifiable, transparent, reproducible and under meaningful human control.

## Introduction

The advent of increasingly autonomous AI systems has fundamentally shifted the central question of AI research. Where earlier generations asked, "can machines think?", the contemporary challenge have become, "can machines by governed, trusted and aligned with human values?". This transformation reflects a growing recognition that technical capabilities of a system alone are insufficient. AI systems operating in sensitive environments must demonstrate verifiable trustworthiness throughout their operational lifecycle.

Current LLM-based systems exhibit persistent challenges of trustworthiness that undermine institutional confidence. Lack of transparency in decision-making processes prevents meaningful oversight, while model drift introduces unpredictable behavioural changes over time. Moreover, non-reproducibility of outputs complicates verification and audit; consequently, leaving stakeholders unable to comprehend, contest, or correct system behaviours. These limitations represent fundamental barriers to responsible deployment in domains where errors carry substantial weight, negatively impacting public safety and financial stability. Furthermore, as agentic AI approaches promise enhanced capabilities through AI-to-AI coordination, they typically focus inward on inter-agent functionality without adequate attention to the governance structures necessary for safe downstream consumption.

As a result of above-mentioned challenges, we have AI deployments with widen accountability gap leading to a divergence between what AI systems can do and what institutions can verify, explain and control. This gap cannot be bridged through ethical guidelines or post-hoc auditing alone. It requires architectural solutions that embed governance directly into the computational fabric of the system; thus, presenting an opportunity to contribute. The proposed Orchestration AI Framework presented in this paper is response to this opportunity.

The Orchestration AI Framework represents a distinct paradigm that extends beyond conventional agentic approaches by integrating humans, AI modules, and information systems into a unified, continuously supervised ecosystem. Through a central Control-Plane, the framework maintains persistent oversight of all interactions; thus, enforcing policies, validating semantic exchanges, and anchoring provenance records in real time.

This paper presents the Ten Criteria for Trustworthy Orchestration AI, a comprehensive assurance framework that operationalises trustworthiness as a set of interdependent, verifiable system components. Specifically, this research aims to: (1) articulate a coherent set of criteria that collectively define Trustworthy AI Orchestration; (2) demonstrate how these criteria can be implemented as runtime properties rather than external controls; and (3) provide a foundation for compliance assessment aligned with international standards and national AI governance frameworks.

## Background and Related Work

The governance of artificial intelligence has attracted substantial attention from standards bodies, regulatory authorities, and research communities worldwide. While substantial attention has been directed toward AI trustworthiness by standards bodies and regulatory authorities, significant gaps remain regarding the application of these principles to orchestrated, multi-component systems.

At the international level, the ISO/IEC standards family establishes the baseline for organisational accountability. The most relevant standard ISO/IEC 38507:2022 (International Organization for Standardization [ISO] provides essential guidance for governing bodies on enabling organisational AI use, establishing higher-level accountability and governance principles for decision-making, data usage, and risk management. As part of the ISO/IEC 38500 series, it extends established IT governance frameworks to address AI-specific considerations including ethical use, compliance, and stakeholder expectations. However, the standard operates at the management level without prescribing architectural properties that AI systems must exhibit, particularly for orchestrated, multi-component ecosystems where governance must be enforced as a runtime property rather than an oversight.

Similarly, International Electrotechnical Commission ([IEC], 2022) and ISO/IEC 42001:2023 provide essential guidance on governance implications and management systems, emphasising board-level responsibility and risk assessment. On the other hand, ISO/IEC 23894:2023 complements these structured approaches for identifying and treating AI-related risks. However, while these standards establish critical organisational procedures, they operate primarily at the management system level and do not prescribe the specific architectural properties required for technical implementation.

On the other side, ISO group has also understood the technical aspects of AI implementations; thus, complementary standards with technical perspective have emerged. For example, the ISO/IEC 5259 (ISO/IEC 5259-1:2024) series (2024-2025) establishes comprehensive frameworks for data quality in analytics and machine learning; ISO/IEC 25059:2023 (soon to be replaced by ISO/IEC DIS 25059) extends the SQuaRE methodology to define quality models for AI systems addressing learning capability, probabilistic outputs, explainability, and fairness; and ISO/IEC TR 24029 provides methods for assessing neural network robustness under perturbed or adversarial conditions. Yet these technical standards address isolated quality aspects without specifying how such properties integrate, collaborate, or are enforced across multi-component systems.

In summary, neither governance nor technical standards address orchestrated AI ecosystems where multiple agents interact under centralised coordination and where governance must be enforced as a runtime property rather than external oversight, plus where data quality, system quality, and robustness guarantees must be maintained across component boundaries throughout the operational lifecycle. Nevertheless, ISO recognises this gap, and development of new standards specific to multi-agent systems, trustworthiness in complex AI, and agentic protocols is an ongoing initiative.

National and regional frameworks offer more specific guidance but face similar limitations regarding architectural specification. For example, In the United States, the NIST AI Risk Management Framework (AI RMF 1.0) (National Institute of Standards and Technology, 2023) creates a socio-technical structure around the functions of Govern, Map, Measure, and Manage, detailing characteristics such as validity, accountability, and explainability. Yet, as a technology-neutral and voluntary instrument, it offers principles rather than the concrete specifications necessary for orchestration. Similarly, the European Union's Artificial Intelligence Act (Regulation (EU) 2024/1689, 2024) creates a robust risk-based regulatory environment with mandatory conformity assessments for high-risk systems. While the Act significantly advances the regulatory landscape, its requirements predominantly focus on individual AI systems rather than the complex, interacting ecosystems found in orchestration. Within Australia, the National Framework for the Assurance of Artificial Intelligence in Government (Australian Government, 2024) and the Voluntary AI Safety Standard (Department of Industry, Science and Resources, 2024) align with these international efforts, emphasising inclusion of Australia's AI Ethics Principles (Department of Industry, Science and Resources, 2019), transparency, accountability, and human oversight, yet they too operate at the policy level without dictating the technical architectures required to achieve these objectives.

Although it may appear that AI governance initiatives are undertaken as solo efforts by individual governments, there are significant intergovernmental efforts such as the OECD AI Principles (OECD, 2024) that establish shared foundations for trustworthy AI. Adopted in 2019 and updated in 2024, the OECD AI Principles represent the first intergovernmental standard on artificial intelligence, with 47 adherents including the European Union and G20 nations. The principles establish five values-based pillars: (1) inclusive growth and well-being; (2) respect for human rights and democratic values; (3) transparency and explainability; (4) robustness, security, and safety; and (5) accountability, complemented with five recommendations for governments on investment, digital ecosystems, enabling policy environments, human capacity building, and international co-operation. This intergovernmental consensus is uniquely significant because it transcends individual national frameworks to establish a global reference point for trustworthy AI governance, with definitions of AI systems and lifecycles now adopted by jurisdictions worldwide including the European Union, United States, and Japan. However, whilst the OECD Principles articulate the "why" and "what" of trustworthy AI, they do not prescribe the "how", leaving organisations without concrete architectural mechanisms to operationalise these values at runtime. This presents an excellent opportunity for contribution: by demonstrating how the Ten Criteria framework translates OECD's intergovernmental consensus into verifiable system behaviours, our work bridges the gap between policy aspiration and technical implementation, offering a pathway for jurisdictions worldwide to achieve architectural alignment with these globally recognised principles.

In parallel academia and research domain has also contributed to the domain of AI Governance. In the last five years, as the applied and agentic AI reaches to average user, several studies have made meaningful impact. For example, the foundational work done by Janssen et al. (2020) have focused on the necessity of data governance and organisational oversight for trustworthy AI. They primarily target the pre-processing data layer. However, our work applies governance into the runtime, translating organisational policies into an architectural Control-Plane that actively orchestrates interactions between multiple AI components. Kuziemski and Misuraca (2020) provide valuable empirical analysis of automated decision-making in public sector contexts, examining case studies from Canada, Poland, and Finland to illuminate how AI deployments can intensify power asymmetries between states and citizens. Their identification of "abstraction traps" and multi-level barriers offers important diagnostic insights. However, their work remains primarily analytical, identifying problems without providing architectural solutions. The Ten Criteria framework advances beyond this diagnostic contribution by specifying implementable architectural properties including Control-Plane coordination, semantic communication integrity, and cryptographic provenance which transform governance requirements from policy aspirations into engineered system behaviours. Where Kuziemski and Misuraca observe that human oversight frequently fails in practice, our framework mandates human governance as a computational property enforced at runtime. Similarly, Wirtz et al. (2020) provide a compelling theoretical basis for AI governance, framing AI risks as 'market failures' that necessitate regulatory intervention through a structured policy cycle. However, they acknowledge that such external regulatory processes suffer from inherent latency. Our framework complements this policy-centric view by internalising regulation into the system architecture. We replace the slow feedback loops of bureaucratic evaluation with the real-time enforcement capabilities of the Control-Plane, ensuring that the 'market failures' Wirtz identifies are pre-empted at the architectural level.

Parallel to these regulatory efforts, the IEEE P7000 series addresses ethical considerations within the engineering lifecycle, with specific standards targeting transparency (P7001), data privacy (P7002), and algorithmic bias (P7003). While these standards provide essential guidance for ethical system design, they function primarily as process frameworks. They do not prescribe the runtime architectural mechanisms necessary to orchestrate and govern interactions between multiple AI components in a live environment.

Although we captured novel and recent works; however, there is substantially more body of work in the domain then mentioned above. Despite these contributions, critical gaps persist:

1. Existing frameworks typically address individual AI systems rather than orchestrated ecosystems comprising multiple interacting components.
2. Governance requirements are frequently specified at abstract levels without corresponding architectural implementations.
3. The interdependencies among trustworthiness properties, such as, how transparency enables accountability, how semantic integrity supports explainability, how provenance underpins auditability, are seldom articulated systematically.
4. Very few frameworks address the full lifecycle of AI systems from design through deployment to decommissioning within a unified governance structure.
5. The epistemological foundations of trustworthy AI, that is, what it means for a system to "know" its limitations, to reason soundly, and to update beliefs appropriately, remain underexplored in governance frameworks.

The limitations identified in existing frameworks motivated the development of a comprehensive approach that addresses orchestrated AI systems holistically. Three foundational insights shape this work.

Firstly, trustworthiness in AI orchestration cannot be achieved through external oversight alone; it must be engineered into system architecture as a first-class concern. Just as safety-critical systems in aviation and nuclear power embed safety mechanisms into their fundamental design, trustworthy AI orchestration requires governance capabilities woven into the computational fabric. This insight motivates the central role of the Control-Plane in our framework which is a coordination layer that enforces policies, validates communications, mediates knowledge, and anchors provenance records as intrinsic system functions rather than afterthoughts.

Second, the criteria for trustworthy orchestration are fundamentally interdependent. Human governance requires policy enforcement to be effective; policy enforcement requires semantic integrity to be meaningful; semantic integrity requires symbolic mediation to be verifiable; and all criteria require provenance anchoring to be auditable. This recognition of interdependence motivates the organisation of criteria into coherent categories while emphasising their cross-cutting relationships within the unified Control-Plane architecture.

Third, trustworthy AI orchestration must be grounded in sound epistemological principles. The framework draws upon established philosophical traditions: Popper's critical rationalism informs the commitment to falsifiable claims and structured argumentation; Piaget's constructivism guides the approach to incremental knowledge evolution; and Clancey's situated cognition illuminates the relationship between symbolic reasoning and contextual action. These foundations ensure that the framework addresses not merely what AI systems do, but how they represent, reason about, and update their knowledge in ways amenable to human understanding and oversight.

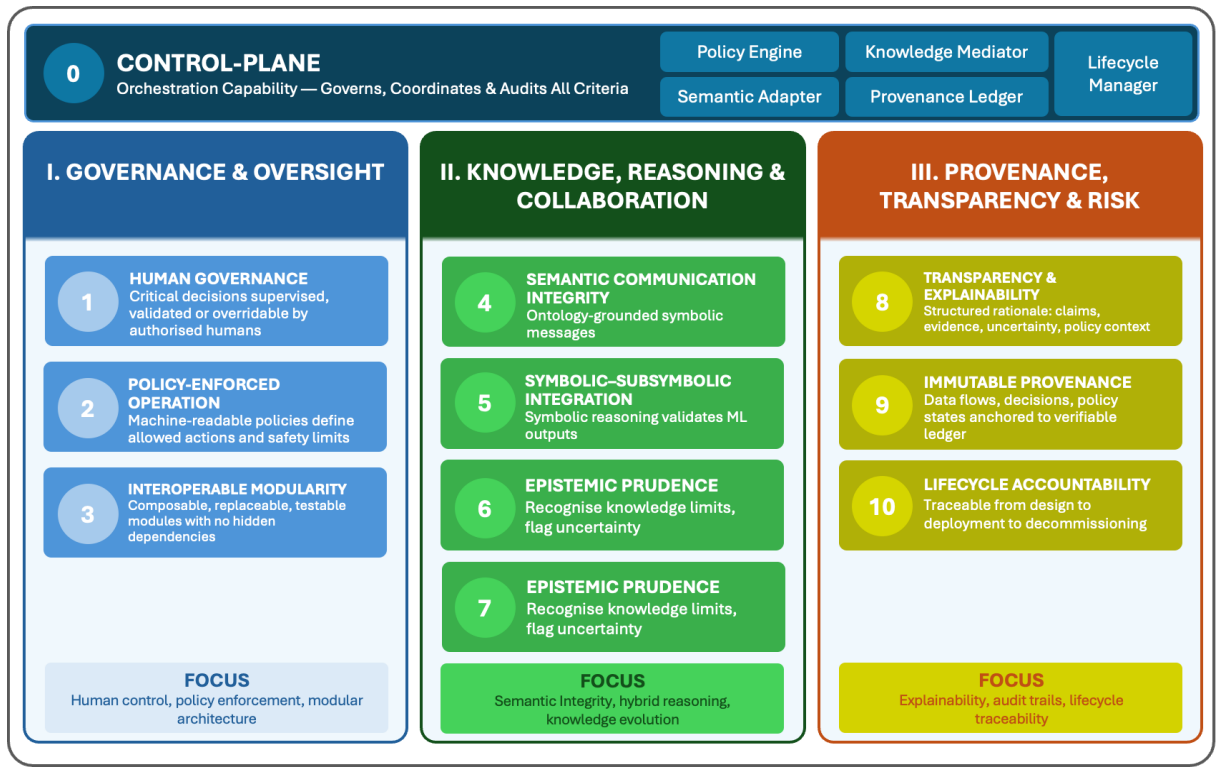

Fig.1. Trustworthy Orchestration AI Framework Overview

Illustrated in figure 1, the Ten Criteria for Trustworthy Orchestration AI Framework are organised into three functional categories overseen by a central Control-Plane. Governance and Oversight encompasses Human Governance (Criterion 1), Policy-Enforced Operation (Criterion 2), and Interoperable Modularity (Criterion 3) which collectively ensures that human authority is preserved, policies are enforced computationally, and system components remain independently verifiable. Knowledge, Reasoning, and Collaboration comprises Semantic Communication Integrity (Criterion 4), Symbolic–Subsymbolic Integration (Criterion 5), Epistemic Prudence (Criterion 6), and Incremental Knowledge Evolution (Criterion 7) which addresses how information flows between components, how different reasoning paradigms are reconciled, how uncertainty is recognised and escalated, and how knowledge is updated safely. Provenance, Transparency, and Risk include Transparency and Explainability (Criterion 8), Immutable Provenance and Observability (Criterion 9), and Lifecycle Accountability (Criterion 10) which ensures that system behaviours can be explained, that records are tamper-evident and complete, and that accountability persists across the full system lifecycle. Each criterion is explained in detail in the following sub-section.

This paper makes the following contributions to the field of trustworthy AI:

- **A Comprehensive Assurance Framework:** We present ten interdependent criteria that collectively define trustworthy orchestration, organised into three functional categories and unified by a Control-Plane architecture. Unlike existing frameworks that address individual aspects of trustworthiness in isolation, our framework treats these properties as an integrated system.
- **Architectural Operationalisation:** We demonstrate how abstract governance requirements can be translated into concrete architectural properties. Each criterion is specified with implementation mechanisms, enabling practitioners to move from principles to deployable systems.
- **Standards Alignment:** The framework is designed for compatibility with ISO/IEC 38507:2022, ISO/IEC 42001:2023, ISO/IEC 23894:2023, the NIST AI RMF, the EU AI Act, and National framework for the assurance of artificial intelligence in government. This alignment facilitates adoption within existing regulatory and compliance structures.
- **Epistemological Grounding:** By explicitly connecting criteria to established philosophical traditions in epistemology and cognitive science, we provide principled foundations for understanding why these criteria are necessary and how they relate to fundamental questions of knowledge, reasoning, and justified belief.
- **Lifecycle Perspective:** The framework addresses trustworthiness across the complete system lifecycle; from initial design through operational deployment to eventual decommissioning, ensuring that accountability is maintained as systems evolve.
- **Domain Applicability:** Through detailed examples spanning clinical decision support, financial services, industrial automation, and multi-institutional coordination, we demonstrate the framework's applicability across high-stakes domains where trustworthy AI orchestration is essential.

# Ten Criteria for Trustworthy Orchestration

The criteria begin with a control-plane layer that sits above everything else. This layer guides and coordinates all other criteria. It keeps the workflow steady, makes sure each part follows the same rules, and supports clear flow across all modules.

The first group, Governance and Oversight, sets the basic rules for safe operation. It covers human control, policy control, and modular design. These criteria make sure people stay in charge, actions follow clear policies, and modules can be checked, replaced, and tested without hidden links.

The second group, Knowledge, Reasoning, and Collaboration, focuses on how information moves and how decisions are made. It covers structured messages, links between symbolic and neural steps, checks on uncertainty, and safe ways to update knowledge. These rules help the system think in a clear way, talk in a stable way, and grow in a controlled way.

The third group, Provenance, Transparency, and Risk, makes all actions visible and traceable. It covers clear explanations, full records of data and decisions, and tracking of models, rules, and policies across their whole life. These criteria help keep trust, reduce risk, and support strong oversight at every step.

| Category | Criterion Number | Criterion Name | Criterion Statement |
| --- | --- | --- | --- |
| Orchestration Fabric | 0 | Control-Plane (Orchestration Capability) | A central control layer guides, checks, and coordinates all other criteria. |
| I. Governance & Oversight | 1 | Human Governance | All critical decisions must be supervised, validated, or overridable by authorised human operators. |
| | 2 | Policy-Enforced Operation | All AI modules and workflows must run under clear, machine-readable policies that define allowed actions and safety limits. |
| | 3 | Interoperable Modularity | AI components must work as composable, replaceable, and independently testable modules with no hidden dependencies. |
| II. Knowledge, Reasoning & Collaboration | 4 | Semantic Communication Integrity | All module interactions must use structured, validated, and ontology-grounded symbolic messages. |
| | 5 | Symbolic–Subsymbolic Integration | Symbolic reasoning must mediate and validate outputs from subsymbolic models. |
| | 6 | Epistemic Prudence | The system must recognise its knowledge limits and flag uncertain or high-risk cases for human review. |
| | 7 | Incremental Knowledge Evolution | Learning and updates must follow governed, versioned, and reversible update processes. |
| III. Provenance, Transparency & Risk | 8 | Transparency & Explainability | Each system action must provide a structured rationale that shows claims, evidence, uncertainty, and policy context. |
| | 9 | Immutable Provenance & Observability | All data flows, decisions, and policy states must be recorded and anchored to a verifiable ledger. |
| | 10 | Lifecycle Accountability | All models, rules, and policies must be traceable from design to deployment to decommissioning. |

Table 1. Trustworthy Orchestration AI by the 10 Criteria (with Control-Plane)

## Criterion 1: Human Governance

### Definition

Human governance means the system cannot complete any high-risk action without human control. A human must supervise, check, or stop the action when needed. The system must support override at any time by an authorised operator. *Dual-control and human-in-the-loop (HITL) mechanisms are mandatory for high-impact actions*

### Purpose

The purpose is to ensure humans remain responsible for consequential outcomes and to preserve human agency in high-stakes decisions. Human governance creates accountability, provides a safeguard against model errors and corrupted data, and enables rapid correction when systems operate outside intended parameters.

*Key Control Mechanisms*
- **Mandatory review gates:** High-risk actions route through HITL stages before execution
- **Override capabilities:** Authorised operators can halt, modify, or reject system recommendations at any time
- **Dual-control workflows:** Actions exceeding risk thresholds require approval from multiple qualified actors
- **Model reasoning exposure:** Systems must expose input features, reasoning traces, and alternative options to human reviewers during HITL gates
- **Human action logging:** Complete records of human decisions, timestamps, and operator credentials

### Example

A clinical decision-support agent processes patient records, laboratory results, and medication history to recommend adjustments to high-risk drugs. When the system flags a medication change as safety-critical, the workflow enters a mandatory human-in-the-loop stage. A licensed clinician opens the validation interface, reviews the system's input features and reasoning trace, and examines lab values using standard references.

The system blocks all downstream execution, no orders are generated, no pharmacy messages are sent until the clinician provides explicit written approval. For medications with narrow therapeutic windows, the system automatically escalates to dual control: a second clinician must independently review and co-sign the action.

Throughout this process, the orchestration engine logs each action, timestamp, and operator ID. This structure ensures the system supports clinical reasoning while guaranteeing that treatment modifications remain under human authority and expert judgment.

## Criterion 2: Policy-Enforced Operation

### Definition

Policy-enforced operation requires that every AI module and workflow execute within formally defined, machine-readable policy constraints continuously monitored and enforced at runtime. Policies explicitly define permissible actions, operational boundaries, resource limits, safety thresholds, and ethical constraints. *The Policy & Governance Engine enforces these rules in real time*

### Purpose

The purpose is to prevent harmful actions proactively through executable constraints that scale uniformly across interactions. It ensures consistent boundary enforcement, encodes regulatory requirements directly into system logic, and enables measurable compliance auditing.

*Key Control Mechanisms*
- **Policy definition and versioning:** Formal syntax for expressing constraints (attribute-based access control, temporal limits, resource quotas)
- **Real-time constraint validation:** Every action request evaluated against active policies before execution
- **Boundary enforcement:** Hard limits on critical parameters (e.g., transaction amounts, approved data types, geographic restrictions)
- **Automatic denial and redirection:** Out-of-policy requests rejected, logged, and routed to escalation paths
- **Policy audit trails:** Complete records of active policies, violation attempts, and enforcement actions

- **Dynamic policy updates:** Policies modified and deployed without system restarts
- **Formal exception mechanism:** Time-limited, pre-approved policy exceptions logged and auditable

### Example

A financial services orchestration system processes automated payments and fund transfers. The Policy & Governance Engine maintains active policies:

- **Transaction boundaries:** Individual transactions <=$500,000; daily limits = 150% of historical average; international transfers to pre-approved countries only
- **Access permissions:** Accounts-payable module can only initiate payments to approved vendors; treasury module can execute wire transfers; operations module cannot
- **Data access constraints:** Personnel records accessible only by HR-authorised modules; customer financial data cannot merge with marketing analytics
- **Temporal rules:** Transfers execute only 9AM–5PM EST; high-risk transactions require review delays
- **Resource consumption:** No workflow exceeds 20% GPU resources or 5-minute execution time

When an AI agent requests a $2.3M payment to an unapproved jurisdiction, the engine detects three violations (amount exceeds threshold, unauthorised destination, requesting module lacks authority). It rejects the transaction, logs the attempt, and routes it to escalation. Regulatory compliance boundaries stay fixed, and policy enforcement sets firm operational limits for the system.

## Criterion 3: Interoperable Modularity

### Definition

Interoperable modularity means each AI component operates as a composable, replaceable, and independently testable module inside the orchestration ecosystem. Every module exposes clear interfaces, follows shared communication standards, and declares all required inputs and outputs.
*No hidden dependencies; every capability is registered and governed*

### Purpose

The purpose is to keep the system robust, maintainable, and safe to extend. It prevents cascading failures by isolating faults, and it allows the system to update or substitute individual components without affecting others. This keeps the orchestration ecosystem stable even when new algorithms, tools, or services are introduced.

*Key Control Mechanisms*
- **Explicit API contracts:** Each module defines its operations, payloads, and response guarantees
- **Central coordination:** Orchestration layer manages all cross-module communication
- **Module isolation:** No shared state, hidden channels, or side effects across components
- **Independent test harnesses:** Each module validated alone and inside composed workflows
- **Versioned module deployment:** Each module versioned independently; updates do not require system-wide restarts
- **Local fault containment:** Module faults detected, logged, and surfaced to orchestrator without cascading to other modules

## Example

A hospital orchestration environment includes a vital-signs anomaly detector, a clinical summarisation agent, and a workflow engine for care-team alerts. Each module registers its interface and expected payload formats. The anomaly detector outputs structured risk events but does not call the summarisation agent directly; the orchestrator routes the event.

When a new summarisation agent is deployed with improved temporal modelling, the module is swapped in through a configuration update without modifying other modules. The orchestrator continues routing calls because the interface contract stays unchanged. Any issues in the new module are isolated and do not affect the anomaly detector or alert engine.

This structure keeps the clinical platform stable even when individual modules evolve or are replaced.

## Criterion 4: Semantic Communication Integrity

### Definition

Semantic communication integrity means all module interactions occur through structured, validated, and ontology-grounded symbolic messages. Each message must follow shared vocabularies, controlled terminology sets, and strict data schemas so that meaning stays stable as information moves across modules.
*This guarantees that meaning, uncertainty, and provenance remain interpretable and auditable across the system.*

## Purpose

The purpose is to prevent misunderstandings between modules caused by different naming conventions, mismatched units, or inconsistent interpretations. Semantic integrity ensures the system produces stable outcomes because each module receives information in a predictable, clearly defined form.

*Key Control Mechanisms*

- **Shared ontologies and vocabularies:** All modules use agreed terminology sets and controlled dictionaries
- **Schema enforcement:** Input and output messages validated against strict structural and semantic schemas
- **Semantic versioning:** Changes to meaning or structure follow clear version control rules
- **Consistency checks:** Runtime validation ensures messages keep the same meaning across modules
- **No silent transformations:** Any data conversion or mapping must be explicit, logged, and governed
- **Runtime message validation:** System checks every message conforms to schema and ontology before routing to target module

## Example

A healthcare orchestration system integrates a clinical decision-support module, a pharmacy ordering module, and a laboratory results module. Without semantic integrity, communication failures occur: the clinical module recommends "Metformin 500," the pharmacy module interprets this as milligrams but records it as micrograms, and the laboratory module tracks glucose levels in different units (mg/dL vs mmol/L), leading to dosing errors and conflicting clinical records.

With semantic communication integrity, the system enforces a shared ontology:

*Shared Ontology and Controlled Vocabularies:*

- Medication names mapped to RxNorm codes (Metformin = RxNorm 6809)
- Dosage must specify numeric value + standardised unit (e.g., "500 mg," not "500")
- Laboratory values must include measurement code (LOINC), numeric result, standardised unit (e.g., glucose = LOINC 2345–4, value = 120, unit = "mg/dL")
- Temporal references must use ISO 8601 format (2024-11-24T14:30:00Z)

When the clinical module sends a medication request, it generates a structured message containing: messageID, timestamp, source and target modules, action type, medication RxNorm code, dosage value with standardised unit, and confidence score. The orchestration layer validates this message against the schema. If any required field is missing or uses non-standard format, the message is rejected.

The dispensing module receives the validated message, parses it using the same ontology, and processes the exact request without ambiguity. The inventory module updates stock counts using the same unit conventions. This eliminates dosing errors and ensures all modules maintain synchronised, interpretable records.

## Criterion 5: Symbolic–Subsymbolic Integration

### Definition

Symbolic reasoning layers (RDR, CBR, KG, logic) must mediate and validate outputs from subsymbolic models, ensuring explainability and epistemic control. These symbolic layers check that model behaviour stays aligned with explicit human knowledge, domain rules, and safety constraints.
*This enables data-driven models to remain interpretable, constrained, and aligned with explicit human knowledge.*

### Purpose

The purpose is to ensure that subsymbolic models do not act as uncontrolled black boxes. Symbolic layers give structure, constraints, and meaning to model behaviour. They help stop unsafe or logically inconsistent outputs and make it easier to trace how conclusions follow from evidence. This approach allows the system to combine the flexibility of machine learning with the clarity and reliability of human knowledge.

*Key Control Mechanisms*

- **Rule-based validation:** Symbolic rules check that model outputs do not violate domain constraints
- **Knowledge-graph grounding:** Model predictions mapped to known entities and relations before use
- **Logical consistency checks:** Symbolic reasoning engines detect contradictions or invalid inferences
- **Hybrid confidence scoring:** Subsymbolic predictions combined with symbolic validation scores; final confidence reflects both sources
- **Symbolic constraint enforcement:** Symbolic layers block, revise, or reshape model outputs that exceed safety limits or violate domain rules

### Example

A fraud-detection orchestration system integrates a deep-learning transaction classifier with a rule-based fraud guideline engine and a merchant-risk knowledge graph. When the classifier predicts "high fraud probability" for a transaction, the orchestration layer does not act on this directly.

It sends the output to the symbolic reasoning layer for validation. The knowledge graph checks whether the transaction aligns with known fraud patterns: merchant category, transaction amount, customer history, and velocity. The rule engine applies fraud-prevention constraints, such as "transactions over $5,000 to new merchants require additional scrutiny" or "repeated rapid transactions within 10 minutes indicate account compromise."

If the classifier suggests fraud but the transaction matches legitimate patterns in the knowledge graph and satisfies all guideline constraints, the symbolic layer adjusts the confidence score downward. If the prediction conflicts with the merchant's known business profile, the layer flags the inconsistency.

Through this hybrid pipeline, subsymbolic models provide strong predictive insights while symbolic layers keep fraud decisions logically consistent with established rule.

## Criterion 6: Epistemic Prudence

### Definition

The orchestration system must maintain awareness of the limits of its own knowledge, detecting when reasoning remains valid but semantically uncertain or incomplete, and flagging such cases for human review and governed knowledge evolution.
*This allows the system to acknowledge uncertainty and seek oversight before acting beyond its understanding.*

### Purpose

The purpose is to prevent the system from making confident but unsupported recommendations. Epistemic prudence ensures the system acknowledges gaps and alerts humans when available knowledge is insufficient for safe action. This reduces harm from model overconfidence, incomplete context, or weak evidence signals.

*Key Control Mechanisms*
- **Uncertainty detection:** The system monitors prediction confidence, distribution shifts, out-of-scope inputs, and missing evidence
- **Knowledge-gap identification:** Symbolic layers check whether required rules, concepts, or relations are absent or incomplete
- **Escalation triggers:** When uncertainty crosses a threshold, the system routes the case to human review workflows
- D**egraded operation modes:** System halts automation, switches to safer defaults, or narrows action scope when knowledge is insufficient
- **Semantic mismatch detection:** System flags new patterns that do not align with known ontologies or established clinical/operational pathways

### Example

A clinical orchestration system supports diagnostic reasoning for complex multimorbidity cases. A patient presents with an unusual combination of symptoms:

moderate fever, atypical chest discomfort, normal inflammatory markers, and a rare medication interaction profile. The subsymbolic model generates a low-confidence prediction because the symptom pattern does not match any clusters in its training distribution.

The symbolic reasoning layer checks clinical guidelines and the knowledge graph. It detects several gaps: no clear pathophysiological relation links the symptoms, and the medication interaction is not represented in the current knowledge base. The orchestration system therefore marks the case as epistemically uncertain.

Instead of producing a forced diagnostic suggestion, the system triggers an escalation workflow. It informs the clinician that the current evidence does not support a confident output and presents the factors causing uncertainty: missing interaction rules, weak symptom alignment, and low model confidence. The system then switches to a fallback mode where it provides only safe supportive information, such as relevant differential diagnoses and recommended next diagnostic tests.

At the same time, the case is logged into the governed knowledge-evolution pipeline. Clinical experts later review the flagged pattern, update the medication-interaction rules if needed, and add new symptom relations to the knowledge graph. The next time a similar case appears, the system has stronger and more complete epistemic support.

## Criterion 7: Incremental Knowledge Evolution

### Definition

Learning and adaptation must occur through governed, versioned, and reversible updates. All knowledge changes follow the lifecycle: Propose → Validate → Simulate → Approve → Apply → Anchor → Monitor.
*This structure ensures that changes to rules, ontologies, models, and workflow logic become controlled events rather than uncontrolled drift. Each update remains auditable, traceable, and safe to roll back.*

### Purpose

The purpose is to ensure that knowledge evolves safely without unpredictable or silent changes. It prevents unverified ontology changes and rule drift while allowing the system to incorporate new research and evidence. Every update remains auditable, traceable, and safe to roll back.

*Key Control Mechanisms*
- **Governed update pipeline:** All changes submitted as proposals with clear descriptions and evidence
- **Validation checks:** Clinical experts, domain specialists, or automated reasoning engines confirm correctness

- **Simulation testing:** Updates run in controlled sandboxes to check downstream effects and risk impact
- **Approval workflow:** Authorised reviewers sign off before updates reach operational modules
- **Versioned application:** New knowledge receives version tags, with full rollback paths kept active
- **Anchoring:** Once deployed, updates become fixed reference points for future reasoning and audits

### Example

A clinical knowledge graph encodes disease–symptom relations, comorbidity structures, and evidence-based treatment rules. New peer-reviewed research reports that a rare autoimmune condition presents with a newly identified biomarker. A clinician submits a formal update proposal to add the biomarker node, link it to the disease entity, and adjust the diagnostic rule that uses this biomarker.

The validation team reviews the proposed connection and checks for conflicts with existing laboratory interpretation rules. The simulation engine then applies the updated rule to five years of historical patient cases, confirming that it produces plausible risk estimates without increasing false alarms.

After approval, the orchestrator activates the update as a new version of the diagnostic rule set. Monitoring tools track how often the new biomarker is triggered in real-world cases. If unexpected alert spikes appear, the system pauses the update and reverts to the prior version for safety review.

This controlled process allows the medical knowledge base to evolve with new science while keeping reasoning stable, interpretable, and auditable.

## Criterion 8: Transparency & Explainability

### Definition

Transparency and explainability mean every system action must produce a structured rationale that humans and AI modules can read, interpret, and verify. Each rationale must include the claim made by the system, the evidence used, the uncertainty level, and the relevant policy or rule context.
*This makes the reasoning process visible and contestable, transforming automated decisions into accountable actions.*

### Purpose

The purpose is to ensure that decisions remain understandable and traceable, not opaque or hidden inside model internals. It allows clinicians and operators to see how the system reached a conclusion and whether the reasoning stays aligned with guidelines, data, and domain knowledge. This reduces the risk of silent errors, exposes weak evidence, and supports human review during critical situations. Transparent reasoning also helps with training, regulatory compliance, and post-incident investigation.

*Key Control Mechanisms*

- **Structured rationale objects:** Each output includes claims, supporting evidence, rules used, model features, and uncertainty values
- **Evidence linking:** Outputs linked to data sources, knowledge-graph nodes, guideline references, or provenance trails
- **Uncertainty disclosure:** Confidence values and ambiguity indicators shown in a clear, audited format
- **Explainable model layers:** Symbolic traces, attention maps, or decision paths accessible for review
- **Policy-context annotations:** Action rationales include which policies, constraints, or rules affected the decision
- **Audit-ready exports:** Explanations recorded so investigators can reconstruct past decisions

## Example

A hospital orchestration system detects patient deterioration risk. When a subsymbolic model flags possible sepsis, the system generates a structured rationale rather than a binary alert. The rationale includes:

- **Claim**: "Patient at elevated sepsis risk"
- **Evidence**: Vital-sign trends (heart rate elevation, respiratory rate increase), abnormal laboratory markers (lactate > 2 mmol/L), recent clinical notes mentioning concern for infection
- **Confidence**: Model confidence 0.87 for this patient class
- **Uncertainty**: Two supportive signs present; immune markers conflicting
- **Policy context:** Sepsis screening rules applied (lactate threshold crossed, sustained respiratory rate > 22)
- **Guidelines:** Hospital sepsis protocol version 2.3 (active as of 2024-11-01)

The clinician receives a clear, structured explanation showing why the alert was generated, how strong the evidence is, and which policies influenced the decision. The orchestration engine records this rationale so clinical auditors can later reconstruct the reasoning.

## Criterion 9: Immutable Provenance, Observability & Risk Containment

### Definition

Immutable provenance, observability, and risk containment mean all data flows, system decisions, model outputs, human interventions, and active policy states must be cryptographically anchored to a verifiable ledger. Every event becomes tamper-evident and traceable across the full lifecycle of orchestration.
T*his guarantees that lineage, authorship, uncertainty, and policy context remain auditable even across organisational or regulatory boundaries*.

### Purpose

The purpose is to ensure accountability across the full decision lifecycle, even when actions cross teams or institutions. Immutable provenance makes it possible to reconstruct how decisions were made, which evidence was used, and which policies were active. This reduces legal, clinical, and operational risks by preventing hidden modifications or loss of event records. Observability supports incident response by enabling investigators to replay events and diagnose failures.

*Key Control Mechanisms*

- **Cryptographic event anchoring:** Each action, message, or model output hashed and written to a verifiable ledger
- **End-to-end lineage tracking:** Full trace of data origins, decision steps, policy versions, and human approvals
- **Risk-containment boundaries:** Isolation rules that prevent uncontrolled propagation of faulty outputs across modules
- **Anomaly detection in logs:** Observability tools detect unusual patterns, policy violations, or abnormal system behaviour
- **Immutable audit trails:** Historical states preserved in a form that cannot be deleted or silently changed
- **Policy-state recording:** Governance engine logs which policy version was active during each decision
- **Forensic reconstruction tools:** Ledger-backed evidence allows investigators to replay events and diagnose failures
- **Cross-organisation verification:** Ledger structure and cryptographic binding enable auditors at different organisations to independently verify integrity

### Example

A multi-hospital health network uses an orchestration system to coordinate emergency stroke care. When a patient arrives with suspected stroke symptoms, the system processes imaging, clinical notes, and vital signs and generates a recommended care pathway. Each step produces cryptographically anchored entries on a distributed ledger:

- Model processes imaging → output recorded with confidence scores and raw features
- Clinician reviews recommendation → override decision logged with operator ID and justification
- Patient transferred to second hospital → receiving facility verifies every prior entry (cannot be altered)

If the imaging classifier later shows performance drift, observability tools detect a spike in false-positive alerts. Because the system records which classifier version was active during each decision, investigators can replay logged entries to determine when drift began and how decisions changed. The immutable ledger proves what was decided and when.

This combination of immutable provenance and observability stops hidden errors, enables trusted multi-site audit, and protects safety even when multiple systems and personnel participate in decision flows.

## Criterion 10: Lifecycle Accountability

### Definition

Lifecycle accountability means all models, rules, policies, and orchestration assets must remain traceable from initial design through deployment, maintenance, updates, and final decommissioning. Every artifact must have clear version control, rollback paths, and provenance metadata.
*This ensures that the system always knows what is running, why it exists, when it changed, and who approved it.*

### Purpose

The purpose is to prevent the system from operating on unknown or unmanaged logic. Lifecycle accountability makes sure that outdated, unsafe, or unapproved components cannot stay active. It also allows investigators to recreate historical states, understand how decisions were made at a specific time, and detect when changes introduced new risks. This supports long-term governance because each component remains auditable and tied to responsible owners.

*Key Control Mechanisms*
- **Unified asset registry:** Every model, rule, ontology, and policy recorded with version tags, creation date, and complete change history
- **Design-to-deployment documentation:** Each asset linked to design decisions, training data sources, validation evidence, and approval records
- **Ownership assignment:** Each asset tied to accountable individuals or teams responsible for maintenance and correctness

- **Rollback paths:** Any component can revert to a previous stable version without system disruption; rollback events logged
- **Retirement workflows:** Decommissioned models or rules archived with full metadata, reason for withdrawal, and replacement information
- **Historical state reconstruction:** Tools to replay system behaviour under older model or policy versions for investigation purposes

### Example

A large integrated hospital system manages a sepsis-risk prediction model used across emergency, inpatient, and critical care units. When the model was originally developed, the system stored the training dataset description, feature definitions, model architecture, validation metrics, and approval records. During deployment, the model received a version tag that linked it to the exact policy and guideline context active at the time.

After one-year, new clinical evidence shows that additional laboratory markers improve early detection. A new model version is trained, reviewed by clinical governance teams, validated on retrospective cases, and registered through the lifecycle accountability system. The old model remains available for rollback, and its clinical impact history stays fully accessible.

When an auditor later reviews a case from six months earlier, they can reconstruct the exact state of the system at that time:

- which sepsis model version was running
- which clinical rules were active
- which policy thresholds triggered the alert
- who approved the last update
- what rationale and evidence supported activation

If the new version causes unexpected alert saturation, the orchestrator reverts to the previous model and logs the event. The retired version is archived with explanations and all supporting metadata.

This approach ensures that every model, rule, and policy stays fully accountable through its entire operational life and that changes never compromise traceability or clinical safety.

## Orchestration Fabric Criterion 0 — Control-Plane (Orchestration Capability)

A dedicated orchestration layer must govern, coordinate, and audit all module interactions, enforcing semantic integrity, policy compliance, and provenance across the entire system in real time

*This ensures that every function of the orchestration ecosystem operates under a single, coherent source of governance and technical truth.*

## Purpose

The purpose of the Control-Plane is to transform governance from an external oversight function into an intrinsic architectural property. Without a central coordination layer, orchestrated AI systems risk fragmented policy enforcement, inconsistent semantic interpretation, untracked provenance, and ungoverned inter-module dependencies. The Control-Plane addresses these risks by serving as the computational fabric through which all governance requirements are realised.

*Key Control Mechanisms*

- **Unifies coordination:** Prevents ad-hoc, point-to-point module communications that bypass governance checks
- **Enforces consistency:** Ensures all modules operate under the same active policy version, semantic ontology, and operational constraints
- **Maintains observability:** Provides a single point of visibility into system state, message flows, and decision provenance
- **Enables auditability:** Records all interactions in a format suitable for compliance verification, incident investigation, and regulatory reporting
- **Supports graceful degradation:** Detects module failures and routes workflows through fallback paths without compromising safety guarantees
- **Preserves human authority:** Ensures that escalation pathways, override mechanisms, and human-in-the-loop gates remain accessible regardless of system complexity

## Example

A regional health network operates an orchestration system that coordinates clinical decision support, pharmacy dispensing, laboratory result interpretation, and care-team communication across twelve hospitals. The Control-Plane governs all interactions.

For example, a clinical decision-support module processes a patient's laboratory results and vital signs, generating a recommendation to adjust anticoagulation therapy.

*Control-Plane Workflow:*

- **Message reception and semantic validation:** The clinical module sends a structured recommendation message. The Control-Plane validates the message against the shared clinical ontology (ICD-11 codes, LOINC laboratory references, RxNorm medication identifiers). A malformed or ambiguous message would be rejected at this stage.
- **Symbolic mediation:** The recommendation originated from a neural risk-prediction model. The Control-Plane routes the output through the symbolic validation layer, which checks the prediction against clinical guideline rules and the patient's medication-interaction knowledge graph. The symbolic layer confirms logical consistency and attaches a validation score.

- **Policy evaluation:** The Control-Plane invokes the Policy & Governance Engine. Active policies specify that anticoagulation adjustments require dual clinical review. The engine marks the action as requiring HITL escalation.
- **Human escalation:** The Control-Plane suspends downstream execution and routes the case to the clinical review queue. The system presents the reviewing clinician with the structured rationale: input features, model confidence, symbolic validation score, relevant guideline references, and alternative options. The Control-Plane blocks all pharmacy and ordering actions until explicit clinician approval is recorded.
- **Provenance anchoring:** Before the clinician's approval is acted upon, the Control-Plane anchors a complete provenance record: patient identifier (pseudonymised), input data hashes, model version, policy version, symbolic validation outcome, clinician ID, timestamp, and decision outcome. This record is cryptographically signed and written to the immutable ledger.
- **Downstream routing:** With approval recorded and anchored, the Control-Plane routes the validated, approved recommendation to the pharmacy dispensing module. The message includes provenance references so the receiving module can verify the decision chain.
- **Observability and monitoring:** Throughout this workflow, the Control-Plane emits telemetry to the observability dashboard: message latency, policy evaluation duration, escalation queue depth, and provenance anchoring confirmation. Anomaly detection algorithms monitor these metrics for deviations that might indicate system degradation or attack.
- **Lifecycle recording:** The Control-Plane updates the lifecycle registry to reflect that this model version, policy version, and ontology version were active during this decision. If a future audit requires reconstruction of the decision context, the registry provides the necessary version references.
- **Failure Handling:** If the pharmacy module fails to acknowledge receipt within the timeout window, the Control-Plane activates circuit-breaker logic, isolates the failed module, logs the incident, and routes the order through a fallback pathway (direct clinician notification with manual order entry instructions). At no point does the failure propagate to other modules or compromise the provenance record.

# Summary and Future Directions

This paper presents the *Ten Criteria for Trustworthy Orchestration AI*, a framework designed to transform trustworthiness from an aspirational principle into a verifiable, engineered system property. By addressing challenges such as transparency and accountability through architectural integration, the framework relies on a central Control-Plane (Criterion 0) to coordinate ten interdependent criteria. These are organised into three functional categories: *Governance and Oversight* (Criteria 1–3), which preserves human authority and enforces machine-readable policies; *Knowledge, Reasoning, and Collaboration* (Criteria 4–7), which manages semantic integrity and enforces epistemic humility; and *Provenance, Transparency, and Risk* (Criteria 8–10), which anchors decisions to cryptographically verifiable ledgers for full lifecycle accountability.

Beyond its architectural specifications, the work grounds its engineering choices in epistemological traditions including critical rationalism and situated cognition to ensure a coherent approach to machine reasoning. Simultaneously, it operationalises abstract requirements from major regulatory instruments, such as the EU AI Act and ISO/IEC standards, translating governance principles into concrete technical specifications. This dual alignment demonstrates that trustworthy orchestration can be systematically engineered to operate transparently and under meaningful human control.

Moving from theoretical analysis to operational maturity requires addressing several limitations and opportunities for future research. Key priorities include empirical validation through longitudinal case studies and the development of automated tools to verify compliance, thereby reducing manual assessment burdens. The research agenda also highlights the need for formal methods to mathematically prove criterion satisfaction, optimisation strategies to manage the computational overhead of cryptographic anchoring, and the adaptation of the framework to emerging paradigms such as Large Language Model (LLM) orchestration and decentralised agent ecosystems.

# Declaration

AI tools were used solely to improve readability. All suggested edits were reviewed and verified by the authors to ensure the original meaning and intent were preserved.